\documentclass[journal]{IEEEtran}
\usepackage{amsmath,amssymb,amsfonts}
\usepackage{algorithmic}
\usepackage{graphicx}
\usepackage{epstopdf}
\usepackage{booktabs}
\usepackage{multicol}
\usepackage{multirow}
\usepackage[hyphens]{url}
\usepackage[colorlinks=true, allcolors=blue]{hyperref}
\usepackage[linesnumbered,ruled,vlined]{algorithm2e}
\def\BibTeX{{\rm B\kern-.05em{\sc i\kern-.025em b}\kern-.08em
    T\kern-.1667em\lower.7ex\hbox{E}\kern-.125emX}}
\begin{document}
\title{AeroLite-MDNet: Lightweight Multi-task Deviation Detection Network for UAV Landing} 
\author{Haiping Yang, Huaxing Liu, Wei Wu, Zuohui Chen, and Ning Wu. 
\thanks{This work is supported by the Pioneering and Leading Project of Zhejiang Province under Grant 2025C02042
. \textit{(Corresponding author: Zuohui Chen)}} 
\thanks{Haiping Yang is with the Institute of Cyberspace Security, Zhejiang University of Technology, Hangzhou, 310023, China, also with the Binjiang Institute of Artificial Intelligence, ZJUT, Hangzhou, 310056, China (e-mail: yanghp@zjut.edu.cn).}
\thanks{Huaxing Liu is with the College of Computer Science and Technology, Zhejiang University of Technology, Hangzhou, 310023, China (e-mail: 221122120307@zjut.edu.cn).}
\thanks{Wei Wu and Zuohui Chen are with the College of Geoinformatics, Zhejiang University of Technology, Hangzhou 310023, China, also with the Binjiang Institute of Artificial Intelligence, ZJUT, Hangzhou, 310056, China (e-mail: wuwei@zjut.edu.cn, czuohui@zjut.edu.cn).}
\thanks{Ning Wu is with the Quzhou Southeast Digital Economic Development Institute, Quzhou, 310058, China (e-mail: qzxy\_wn@foxmail.com).}}

\maketitle

\def\sectionautorefname{Section}%
\def\subsectionautorefname{Section}%
\def\subsubsectionautorefname{Section}%
\def\figureautorefname{Fig.}
\def\tableautorefname{Tab.}
\def\algorithmautorefname{Alg.}

\begin{abstract}

Unmanned aerial vehicles (UAVs) are increasingly employed in diverse applications such as land surveying, material transport, and environmental monitoring. Following missions like data collection or inspection, UAVs must land safely at docking stations for storage or recharging, which is an essential requirement for ensuring operational continuity. However, accurate landing remains challenging due to factors like GPS signal interference. To address this issue, we propose a deviation warning system for UAV landings, powered by a novel vision-based model called AeroLite-MDNet. This model integrates a multiscale fusion module for robust cross-scale object detection and incorporates a segmentation branch for efficient orientation estimation. We introduce a new evaluation metric, Average Warning Delay (AWD), to quantify the system's sensitivity to landing deviations. Furthermore, we contribute a new dataset, UAVLandData, which captures real-world landing deviation scenarios to support training and evaluation. Experimental results show that our system achieves an AWD of 0.7 seconds with a deviation detection accuracy of 98.6\%, demonstrating its effectiveness in enhancing UAV landing reliability. Code will be available at \href{https://github.com/ITTTTTI/Maskyolo.git}{https://github.com/AeroLite-MDNet/AeroLite-MDNet.git}

\end{abstract}


\markboth{IEEE Transactions on Circuits and Systems for Video Technology}%
{}

\section{Introduction}
\label{sec:introduction}

\IEEEPARstart{U}{nmanned} aerial vehicles (UAVs), also known as drones, have been widely used in fire detection, geological hazard monitoring, and dangerous behavior monitoring~\cite{ruvzic2021application} for their agility, compactness, and cost-efficiency. To reduce the dependency of UAVs on human labor and skills, UAV nests are widely used to minimize manual operations, allowing the UAVs to perform autonomous monitoring. UAV nests also offer functionalities such as safe parking, charging, data transmission, routine maintenance, repairs, and communication relays~\cite{knitter2024survey}. However, automatic UAV landing is a major challenge in the application of UAV nests. In this work, we propose a landing deviation warning system for UAV landing.\par

\begin{figure}[t]
\centering
\includegraphics[width=0.5\textwidth]{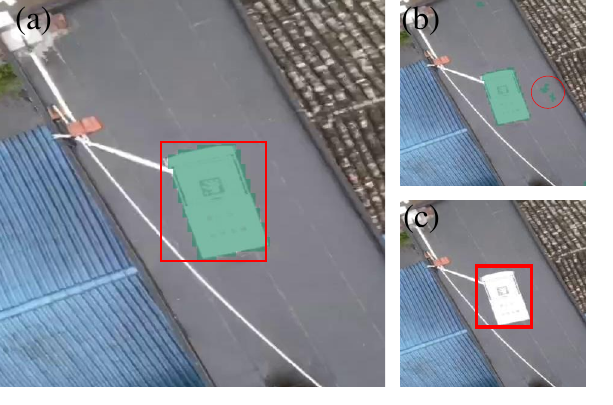}
\setlength{\abovecaptionskip}{-0.3cm}
\caption{Comparison of current methods and ours. (a) The multitask model combining the advantages of segmentation and detection can fit the target area well. (b) The segmentation model has a substantial prediction error (marked in red circle) due to incorrect segmentation. (c) The detection model fails to locate the target area for rotated targets.}
\vspace{-0.4cm}
\label{show}
\end{figure}
Early systems for determining UAV landing deviation primarily relied on global positioning systems (GPS) and ground markers. However, GPS is often inaccurate when estimating ground elevation, particularly in areas with complex terrain or varying topography~\cite{nakamura2017estimation}. Moreover, GPS-based methods are susceptible to interference from geomagnetic storm activities. In densely built urban areas or mountainous regions, the accuracy of GPS can be affected by the presence of obstacles~\cite{wing2005consumer}. With the development of computer vision, vision-based deviation detection has been widely applied in UAVs. Compared to GPS-based methods, vision-based methods offer an efficient and reliable solution for achieving robust autonomous landing. 


The challenge of vision-based UAV landing lies in providing quick and accurate warnings of deviations. Segmentation methods can accurately obtain the target shape, however, due to the lack of detection frame constraints, they may generate artifacts that mislead localization (\autoref{show}(b)) and the model's slow speed makes it challenging to meet real-time application requirements. The target detection methods \cite{zhang2022dino} can locate the target. Nevertheless, the vertical detection frame cannot effectively capture the shape and orientation of the target when it rotates substantially (\autoref{show}(c)). 

To address the issues, we propose a lightweight and multitask learning model, AeroLite-MDNet. We reduce the model parameters by keeping the core architecture and multi-scale fusion unchanged while reducing the depth of the three convolutional layers (C3) module, limiting the parameter size to millions to achieve real-time inference. Data enhancement techniques such as random flipping and image blending are used to improve the generalization to the changes in the environment. Meanwhile, multi-task joint learning is employed to capture multi-level information, where the detection branch provides the location and category of the target, and the segmentation branch provides the shape and region of the target (\autoref{show}(a)).

The main contributions of this paper can be summarized as follows:

\begin{enumerate}

\item For UAV landing deviation warning in real-world environments, we propose AeroLite-MDNet, a lightweight multi-scale detection-segmentation multitask model, representing the first UAV landing deviation detection framework integrating dual-task detection and segmentation.

\item We collect UAVLandData, a landing dataset containing 9,142 images from 13-hour videos across 24 scenarios with diverse weather/lighting conditions, annotated with bounding boxes and pixel-level segmentation masks.

\item  The model achieves 104.2 frames per second (FPS) on RTX 3070 with 7.12M parameters and 98.6\% warning accuracy. Optimized with ONNX acceleration, the model achieves 7.3 FPS on Jetson TX2, demonstrating efficient edge deployment capability.

\end{enumerate}



\section{Related Work}
\label{sec:rela}

This section provides a comprehensive review of research on autonomous UAV landing, with a focus on methods that utilize visual information for positioning. In addition, we discuss related studies on image segmentation and object detection, as these techniques form the foundation of the proposed vision-based warning system.

\subsection{Vision-Based UAV Landing}
Vision-based methods can be divided into model-driven and data-driven methods. Model-driven methods maintain an advantage in interpretability, while data-driven methods have significantly improved adaptability in complex environments. 

\subsubsection{Model-driven Methods}
As the mainstream paradigm of early visual landing research, model-driven methods relied on traditional computer vision techniques, achieving landing localization through manual feature extraction and matching.
Lange et al.~\cite{lange2008autonomous} developed a landing localization method that employed the Hough Transform to identify a concentric landing pattern, thereby determining the location of the landing.
Lee et al.~\cite{6393429} used a morphological algorithm and edge distribution function  to extract features of the international standard helipad.
Fraczek et al.~\cite{fraczek2018embedded} proposed an embedded video subsystem for classifying terrain based on UAV camera images. The system utilized a decision tree, support vector machine classifiers, and a shadow detection module, to determine landing positions.

\subsubsection{Data-driven Methods}
The recent development of deep learning techniques has led to the emergence of end-to-end learning methods for UAVs. For instance, De et al.~\cite{de2019autonomous} trained a multilayer perceptron neural network to recognize artificial markers using fuzzy Mamdani logic. Bicer et al.~\cite{bicer2019vision} designed a network of five convolutional layers to predict heading angles for the autonomous landing. However, the simplicity of the model resulted in low robustness. Jiang et al.~\cite{jiang2023real} proposed the STDC-CT model for real-time segmentation to identify a secure landing site. However, the inference required high-performance GPUs to achieve real-time processing, which can hardly be deployed on CPU-based embedded systems. Rongbin Chen et al.~\cite{chen2024swin} proposed a novel framework based on Swin-YOLOX for vision-based autonomous landing systems, leveraging hierarchical attention mechanisms to achieve precise landing marker identification with a detection accuracy of 98.7\% AP50.


\begin{figure*}[ht]
\centering
\includegraphics[width=\textwidth]{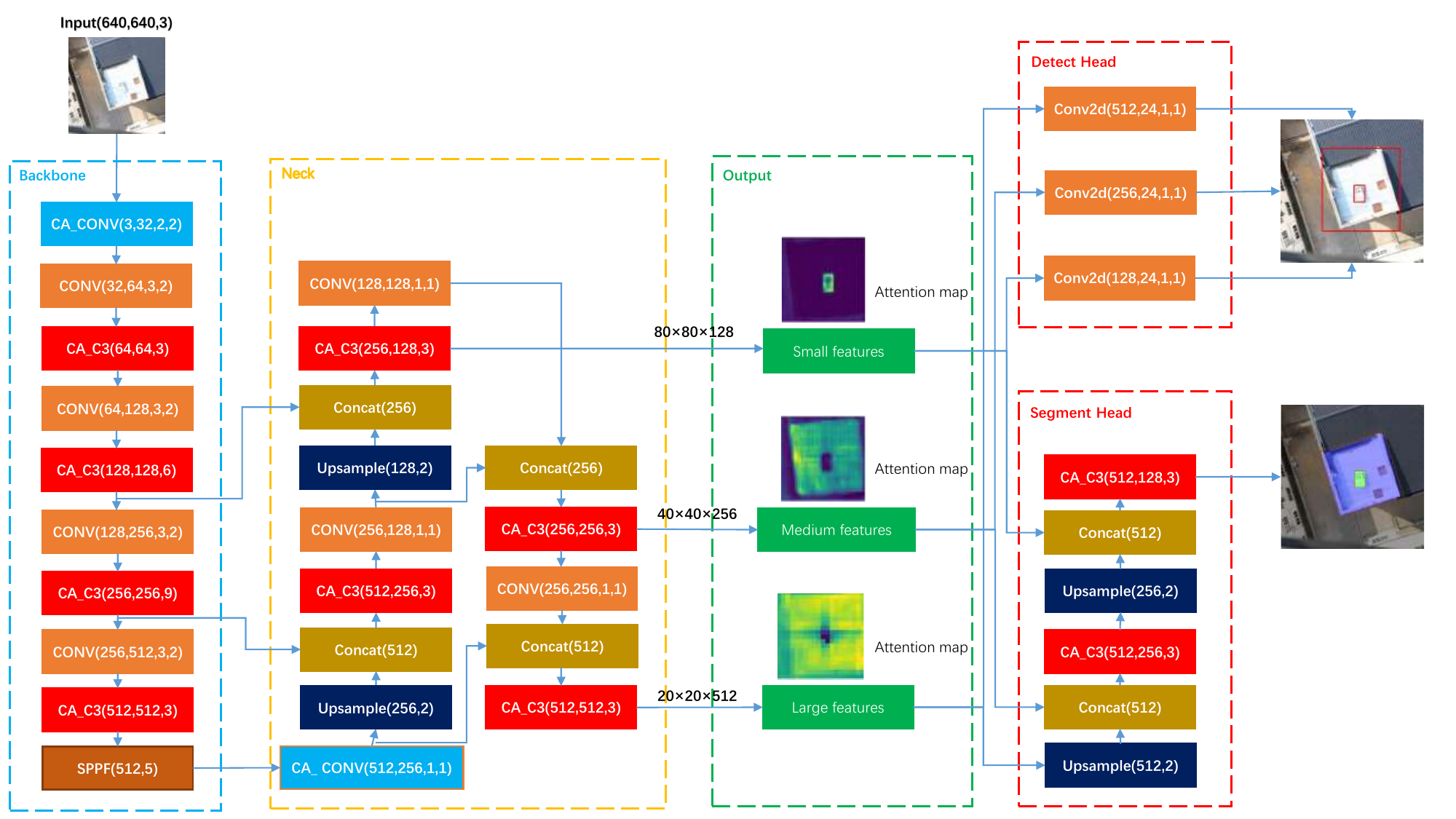}
\setlength{\abovecaptionskip}{-0.3cm}
\caption{
The proposed warning system includes seven steps: (1) during the UAV descent, the video stream is transmitted to the backend; (2) the input video frame is center-cropped to $640\times640$, ensuring the central part of the region of interest is captured; (3) the proposed AeroLite-MDNet model processes video frames;
(4) the multi-task learning part predicts the landing markers through the detection head and segmentation head; (5) the system checks if the target position has any deviation; (6) if there is a deviation, an alert is sent via the internet for manual landing intervention; (7) if no deviation is detected, the next frame is captured and sent into the network for inference.}
\vspace{-0.1cm}
\label{model}
\end{figure*}

\subsection{Object Detection and Segmentation}
Precise identification of landing targets under complex aerial conditions, e.g., scale variation, partial occlusion, requires visual perception techniques. As basic tasks of computer vision, object detection and segmentation algorithms provide critical technical support for UAV landing systems.

Object detection methods are typically categorized into two-stage and single-stage approaches. Two-stage methods, such as Faster R-CNN~\cite{ren2015faster}, utilize a Region Proposal Network to generate candidate regions. The detection model then classifies and localizes the objects within each proposed region. In contrast, single-stage methods like SSD~\cite{liu2016ssd}, Co-DETR~\cite{zong2023detrs}, and YOLO~\cite{jocher2021ultralytics} directly extract features from the input image to predict category and bounding boxes.

In comparison, image segmentation provides pixel-level precision rather than object localization. Long et al.\cite{long2015fully} introduced Fully Convolutional Networks (FCN), which enabled end-to-end training for semantic segmentation tasks. Subsequent methods often employ an encoder-decoder architecture, such as SegNet\cite{badrinarayanan2017segnet} and SSD~\cite{liu2016ssd}. Xie et al.~\cite{xie2021segformer} further advanced the field by proposing SegFormer, which merges the strength of Transformer and CNN, leveraging multi-scale features for more effective segmentation.

Some models have combined object detection and segmentation tasks. He et al.~\cite{He_2017} extended Faster R-CNN~\cite{girshick2015fast} by adding a segmentation branch and proposed Mask R-CNN, a model capable of efficiently performing both detection and segmentation with low additional overhead. Recent state-of-the-art models, such as RTMDet~\cite{lyu2022rtmdet} and Co-DETR~\cite{zong2023detrs}, also integrate both tasks. RTMDet, built on a CNN architecture similar to YOLO, uses CSPDarkNet as its backbone. Co-DETR, a DETR-based network, leverages transformers and enhances learning efficiency through a collaborative hybrid allocation training scheme. 




\section{Methodology}
\label{sec:method}


The workflow of our UAV landing deviation warning system is shown in \autoref{model}. The real-time frames from the video stream are fed into the AeroLite-MDNet. The model comprises four components: 1) feature extraction backbone, 2) neck, 3) Spatial Pyramid Pooling - Fast (SPPF) module, and 4) segmentation and detection head. Downsampling is applied in the backbone module, where global and local features are extracted using the Spatial Channel Attention (SCA) and Channel Attention Conv (CA\_Conv) structures. Subsequently, the neck module processes these features for refinement, generating features at large, medium, and small scales. Additionally, we combine the segmentation head and the detection head for multi-task learning. By the deviation between the expected landing position and the center of the video (the actual landing position), we can determine whether there is a risk of off-course during landing.

\subsection{Backbone and Neck}
\autoref{tab:backbone} gives the overall parameters of the backbone. The architecture comprises nine layers, constructed from four distinct modules, including CA\_Conv, DCNv2, SCA\_C3, and SPPF. The details of each module are described below. 

\begin{table}[!ht]
\centering
\caption{BACKBONE NETWORK PARAMETERS} 
\label{tab:backbone} 
\begin{tabular}{c c c }
    \hline 
    \textbf{Operator} & \textbf{Output} & \textbf{Component}   \\
    \hline
    CA\_Conv & 320×320×32 & kernel:2×2×6, stride:2   \\
    DCNv2 & 160×160×64 & kernel:3×3×128, stride:2  \\
    SCA\_C3 & 160×160×64 & in:128, out:128 \\
    DCNv2 & 80×80×128 & kernel:3×3×256, stride:2  \\
    SCA\_C3 & 80×80×128 & in:256, out:256   \\
    DCNv2 & 40×40×256 & kernel:3×3×512, stride:2   \\
    SCA\_C3 & 40×40×256 & in:512, out:512   \\
    DCNv2 & 20×20×512 & kernel:3×3×1024, stride:2   \\
    SCA\_C3 & 20×20×512 & in:1024, out:1024   \\
    SPPF & 20×20×512 & kernel:5, in:1024   \\
    \hline
\end{tabular}
\end{table}
\subsubsection{CA\_Conv}
\begin{figure}[!t]
\centering
\includegraphics[width=3.5in]{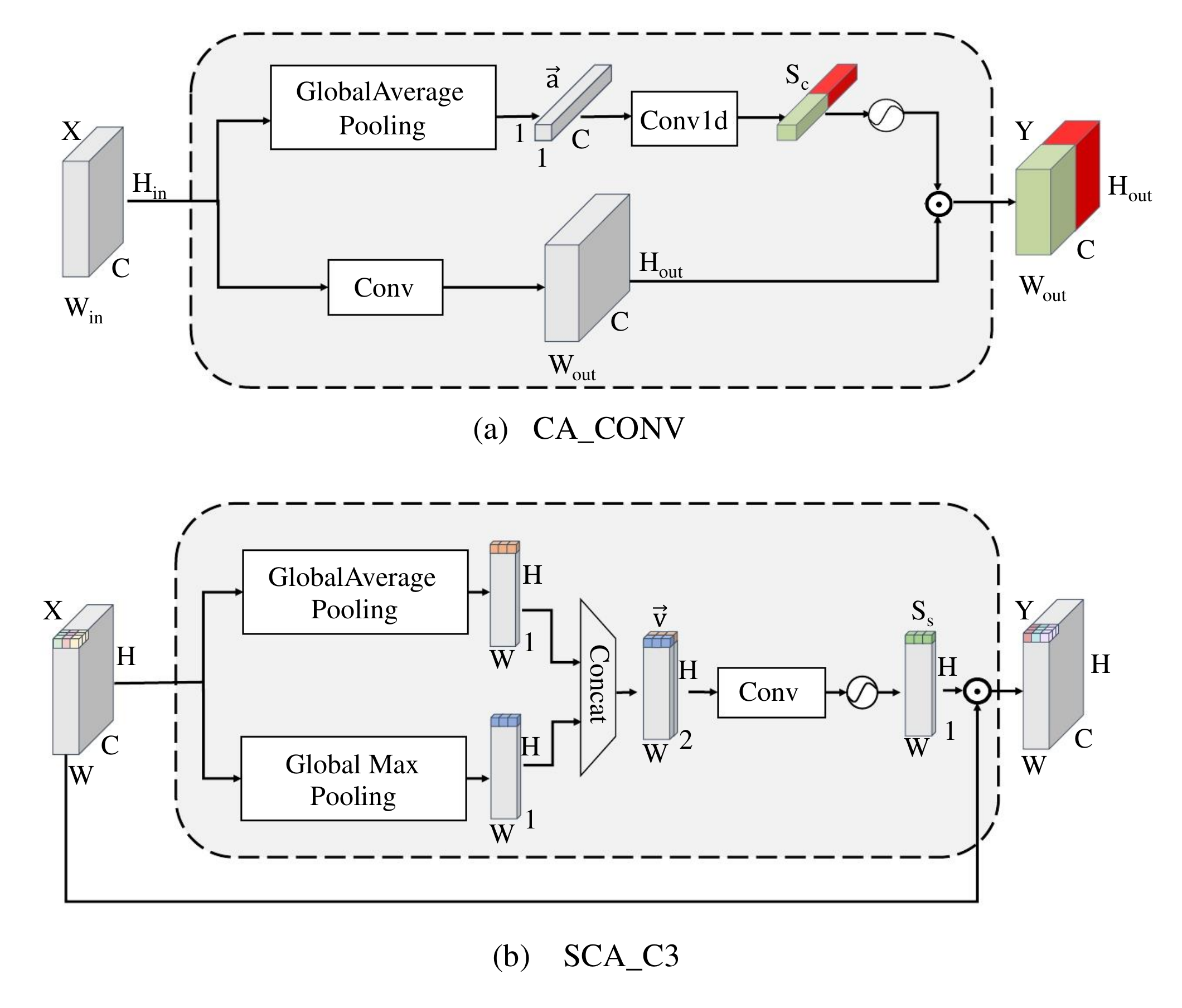}
\setlength{\abovecaptionskip}{-0.3cm}
\caption{Attention modules in the AeroLite-MDNet model, where (a) is the convolutional module based on channel attention mechanism (CA\_Conv). (b) is the Spatial Channel Attention module (SCA). }
\vspace{-0.3cm}
\label{SCA_CA}
\end{figure}

The detailed CA\_Conv modules are shown in \autoref{SCA_CA}.
In the Conv module, we use a $3\times3$ convolutional kernel to keep the original size of the padding and set the step size to 2 instead of using pooling layers to downsample feature maps. This approach reduces the feature map dimension while preserving the key details.


For a given input feature map of dimensions $\{H_{\text{in}},W_{\text{in}}\}$, the size of output $\{H_{\text{out}},W_{\text{out}}\}$ is

\begin{align}
H_{\text{out}} &= \frac{H_{\text{in}}-f_{\text{h}}+2*P}{S}+1 \\
W_{\text{out}} &= \frac{W_{\text{in}}-f_{\text{w}}+2*P}{S}+1,
\label{convw}
\end{align}
where $f_{\text{h}}$ and $f_{\text{w}}$ are the height and width of the convolution kernel, $P$ is the padding size, and $S$ is the stride.
\par

We integrate a channel attention module after convolution to improve its feature extraction ability. Channel attention can identify the most related feature channels and amplify their impact. As shown in \autoref{SCA_CA} (a), the input feature map $X$ first undergoes a global average pooling, wherein the feature values from each channel are averaged, yielding a global feature vector $\vec{a}$
\begin{equation}
\vec{a} = \text{GlobalAveragePooling}(X)
\label{GAP}
\end{equation}
Subsequently, we utilize a linear transformation to obtain the channel attention weights, which are composed of a fully connected layer and a sigmoid activation function.
\begin{equation}
S_{\text{c}} = \sigma(W \cdot \vec{a} + b)
\label{sigmoid}
\end{equation}
where $S_{\text{c}}$ is channel attention weights, $W$ is the weight matrixis, $b$ is the bias, and $\sigma$ is the Sigmoid function. $S_{\text{c}}$ obtained from Equation~\eqref{sigmoid} assigns large values to important channels, enabling the model to capture key features.

In the end, $S_{\text{c}}$ is multiplied by the result of the convolution operation on the input feature $X$, weighting the contributions of each channel.
\begin{equation}
Y =Conv(X) \odot S_{\text{c}}
\label{apply}
\end{equation}
where $Y$ is the weighted output and $\odot$ is element-wise multiplication.\par



\subsubsection{SCA\_C3}
As shown in \autoref{SCA_CA}(b), to enhance the feature extraction capability of the model, we introduce spatial attention and channel attention mechanisms in the C3 module. The C3 module combines different bottleneck layers via channel attention to refine the relevance of different channels to the UAV nest, and then employs the spatial attention mechanism to capture the key spatial features.



This module first applies global max-pooling and global average-pooling to the input feature map $X$ and then concatenates their results to obtain $\vec{v}$.
\begin{equation}
\vec{v} = \text{Concat}(\text{GAP}(X), \text{GMP}(X)).
\label{concat_spetial}
\end{equation}

Subsequently, we reduce the number of channels to 1 by a 2D convolution with a window size of 1. A linear transformation is applied to yield the spatial attention weights $S_{\text{s}}$
\begin{equation}
 S_{\text{s}} = \text{softmax}(\text{conv2d}(\vec{v})).
\label{sigmoid_spetial}
\end{equation}
In the end, the input feature map is multiplied by the spatial attention weights.



\subsection{Head}
Our model has a detection head and a segmentation head, which learn multiple tasks by sharing the feature representations at different scales.


\subsubsection{Detection Head}
The detection head employs preset anchor boxes to identify targets of varying scales. The aspect ratios of these anchor boxes are determined based on statistical analysis of the labeled data. Quick Response (QR) codes are square with an aspect ratio of 1:1, while houses and nests, though their shapes vary depending on the viewing angle, generally exhibit aspect ratios close to 1:2 or 2:1. As a result, the anchor box aspect ratios are set to 1:1 (square), 2:1 (wide rectangle), and 1:2 (tall rectangle), allowing for the detection of objects with different shapes. These anchor boxes help substantially reduce the search space and improve detection efficiency.

To further improve detection across scales, each output feature map generates three sets of anchor boxes corresponding to different object sizes, resulting in a total of nine anchor boxes. For small objects, the feature maps with 8$\times$ downsampling utilize small anchor boxes with sizes \{(10,13), (16,30), (33,23)\}. For 16$\times$ downsampled feature maps, medium anchor boxes with sizes \{(30,61), (62,45), (59,119)\} are used. Similarly, for 32$\times$ downsampled feature maps, large anchor boxes with sizes \{(116,90), (156,198), (373,326)\} are employed to detect large objects.

We set the detection loss as the weighted sum of the classification loss, object loss, and bounding box loss.

{\begin{equation}
\mathcal{L}_{\text{det}} = \frac{3}{N_{l}}*\mathcal{L}_{\text{class}} +\frac{N_{c}}{N_{l}} *\mathcal{L}_{\text{obj}}+\left(\frac{N_{\text{imgsz}}}{N_{\text{model}}}\right)^2 * \frac{3}{N_{l}}* \mathcal{L}_{\text{box}},
\label{eq:detloss}
\end{equation}}
where $N_{l}$ is number of detection layer, $N_{c}$ is number of target detection categories, $N_{\text{imgsz}}$ is the image size, and $N_{\text{model}}$ is the dimension of the model input.
$\mathcal{L}_{\text{obj}}$ and $\mathcal{L}_{\text{class}}$ are focal loss~\cite{lin2017focal}

\begin{equation}
\mathcal{L}_{\text{Focal}}(p_t) = -\alpha_t (1 - p_t)^\gamma \log(p_t),
\label{eq2}
\end{equation}
where, $p_t$ is the predicted probability, $\alpha_t$ is the class weight, and $\gamma$ is the tuning parameter. 

We employ focal loss to address the issue of class imbalance. When a sample is difficult to classify, $p_t$ approaches 1, resulting in a smaller value of $(1 - p_t)^\gamma$, thereby reducing the loss weight for difficult samples. On the other hand, when a sample is easy to classify, $p_t$ is close to 0, resulting in a higher value of $(1 - p_t)^\gamma$, which increases the loss weight for easy samples. This encourages the model to focus on difficult samples.

In Equation~\eqref{eq:detloss}, $\mathcal{L}_{\text{box}}$ is the Complete Intersection over Union~\cite{zheng2021enhancing} (CIoU) loss:

\begin{equation}
\text{CIoU} = \text{IoU} - \frac{{\rho^2(c_2, c_1)}}{{\rho^2(c_2, c_1) + \alpha}} + \text{v}(b_1, b_2),
\label{eq3}
\end{equation}
where $\rho^2(c_2, c_1)$ is the square of the distance between the center points of two boxes $c_2$ and $c_1$, $\alpha$ is a smoothing parameter to prevent zero division when the centers of two boxes coincide, and $\text{v}(b_1, b_2)$ penalizes the deviation between two bounding boxes $b_1$ and $b_2$, by the differences in center distance, width, and height. The penalty term $\text{v}(b_1, b_2)$ is 
\begin{equation}
\text{v}(b_1, b_2) = \frac{\pi}{2} \left( \arccos \left( \frac{\|b_1\| \cdot \|b_2\|}{b_1 \cdot b_2} \right) - \frac{2\pi}{2} \right)^2.
\label{vb1b2}
\end{equation}
We use CIoU to assess the similarity between two bounding boxes, integrating position, size, and overlap.


\subsubsection{Segmentation Head}

\begin{figure}[t]
\centering
\includegraphics[width=0.4\textwidth]{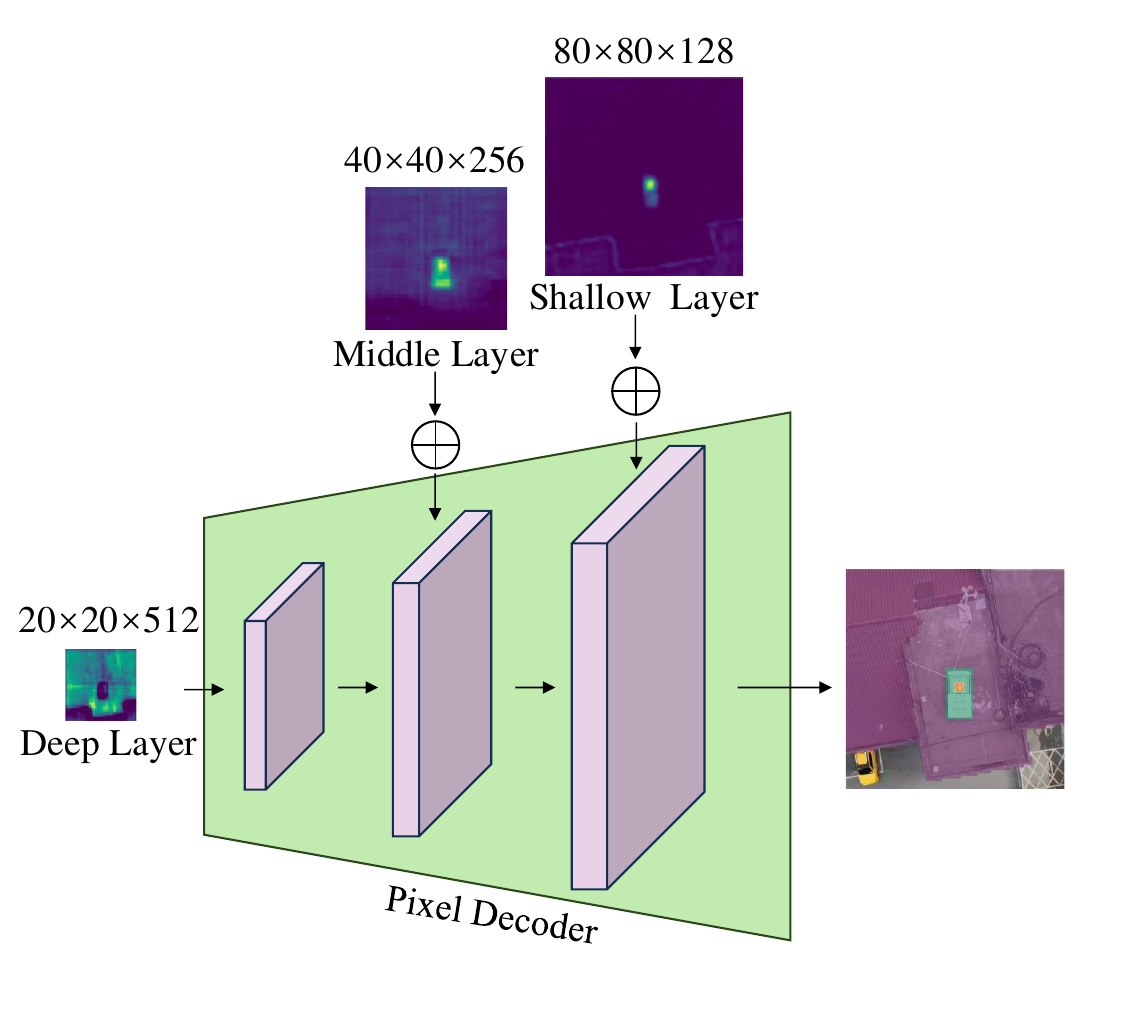}
\caption{Structure of segmentation head. For objects of different scales, we utilize high-resolution features from the neck, passing multi-scale features to the pixel decoder.}
\vspace{-0.1cm}
\label{seghead}
\end{figure}

High-resolution features facilitate identifying objects at different scales, but they are computationally demanding. Therefore, we propose an efficient multi-scale strategy, as shown in \autoref{seghead}, that introduces high-resolution features while managing the computational load. Instead of consistently using high-resolution features, we utilize a feature pyramid that includes both low and high-resolution features.

Specifically, we use a feature pyramid generated by the pixel decoder at resolutions of 1/32, 1/16, and 1/8 of the original image. These embeddings are fed from the lowest to the highest resolution into the corresponding layers of the pixel decoder.

The segmentation loss is the weighted sum of Dice loss~\cite{2016V}, BCEWithLogits loss, and Focal loss. Since we are concerned with the boundaries and overall shapes of the UAV nest and houses,  the weight of the Dice loss is higher than other losses to increase the model's sensitivity to the target regions. 
\begin{equation}
\mathcal{L}_{\text{seg}} = \alpha_1 \mathcal{L}_{\text{Dice}}+\alpha_2 \mathcal{L}_{\text{BCE}}+\alpha_3 \mathcal{L}_{\text{Focal}}
\end{equation}
After experimental testing, the best-performing parameters are $\alpha_1 = 0.5$, $\alpha_2 = 0.25$, and $\alpha_3 = 0.25$.

We use Dice loss to evaluate the performance of image segmentation. Its advantage lies in capturing the overlap between predicted segmentation and ground truth segmentation, making it suitable for addressing edge ambiguity issues. 
\begin{equation}
\mathcal{L}_{\text{Dice}} = 1 - \frac{{2 \times TP}}{{2 \times TP + FP + FN}},
\label{lseg}
\end{equation}
where TP is the number of true positive pixels, FP is the number of false positive pixels, and FN is the number of false negative pixels.

However, the Dice loss focuses on the overall similarity between the predicted and ground truth masks, overlooking individual pixel values. Hence, in scenarios with only foreground and background, even a slight misprediction in small targets can cause a substantial fluctuation in Dice scores. 
To address this limitation, we integrate Dice loss with weighted cross-entropy loss by assigning weights to the small target regions. Additionally, considering the imbalance between classes, we introduce class weights in the loss computation to manage the disparity among different classes as follows:
\begin{align}
\mathcal{L}_{BCE}(y, \hat{y}) = -\frac{1}{N} \sum_{i=1}^{N} \biggl[ \frac{N_{\text{total}}}{2 \cdot N_{\text{positive}}} \cdot y_i \cdot \log(\hat{y}_i) +\\ 
\frac{N_{\text{total}}}{2 \cdot N_{\text{negative}}} \cdot (1 - y_i) \cdot \log(1 - \hat{y}_i) \biggr],\notag
\label{bce}
\end{align}
where $\hat{y}_i$ is the output, , and $y_i$ is the true label.


By combining Dice loss, BCEWithLogits loss, and Focal loss, we can balance both pixel-level similarity and overall classification accuracy. Dice loss measures the overlap of the segmentation, BCEWithLogits loss handles pixel-level classification accuracy, and Focal loss deals with the importance and difficulty of different targets. 
\begin{algorithm}[!ht]
\caption{UAV Landing Decision Process}
\label{alg:UAV_Landing}

\KwIn{Segmentation results of landing video frame $X_{seg}$, predefined threshold $\delta$, vision center$(x_{\text{center}},y_{\text{center}})$, target area $(x_{\text{target}},y_{\text{target}})$, the nest length $l$ in image.}
\KwOut{UAV deviation information}

\BlankLine

\BlankLine
\If{only houses are recognized in $X_{seg}$}{
    \eIf{$(x_{\text{center}},y_{\text{center}})$ in house}{
        \Return{UAV is not deviating\;}
    }{
        \Return{UAV is deviating\;}
    }
}
\If{the nest or the QR code is detected in $X_{seg}$}{
      Calculate offset distance $d$: $d=\sqrt{(x_{\text{center}} - x_{\text{target}})^2 + (y_{\text{center}} - y_{\text{target}})^2}$ \
      
    \eIf{$d/l > \delta$}{
        \Return{UAV is deviating\;}
    }{
        \Return{UAV is not deviating\;}
    }
}

\end{algorithm}
\subsection{Deviation Warning Detection}
Upon completion of the two parallel tasks, the detection branch produces $n$ target boxes and their confidence scores, while the instance segmentation branch generates $n$ masks. The final instance segmentation result is refined by cropping the mask using the detection box, removing the noise outside the bounding box.

After obtaining the segmentation results, we determine deviation according to the algorithm shown in \autoref{alg:UAV_Landing}. In the initial stage of UAV landing, when only houses can be recognized (lines 1-5), whether the UAV deviates depends on whether the house is in the center of the field of view.
Subsequently, the landing area is limited to the upper part of the nest (lines 7-11). We assess whether the distance between the vision center and the landing area exceeded a predefined threshold, which is based on the size of the nest in the image.

\subsection{Metric}
To evaluate the effectiveness of the proposed warning system, we introduce a novel metric called AWD. This metric quantifies the average time difference between the actual onset of deviation and the time at which the system issues a warning.
\begin{equation}
\text{AWD} = \frac{1}{n} \sum_{i=1}^{n} |T_i - P_i|,
\label{AWD}
\end{equation}
where $T_i$ denotes the actual deviation time for the $i$-th landing, $P_i$ represents the system's warning time for the $i$-th landing, and $n$ is the total number of landings. This metric calculates the average time difference between the onset of deviation and the model warning. 

To mitigate the impact of inference delays on the evaluation, frames captured during the model's processing time are excluded. Instead, the last frame captured before inference completes is used. In other words, if the processing time per frame was $k$ seconds, all frames captured during this interval are excluded to ensure consistency. 

    


\section{Experimental Results and Discussion}
\label{sec:exp}

\subsection{Dataset}

\begin{table}[h]
\centering
\caption{DESCRIPTION OF DATASET WITH THE CLASSES: Nest, QRcode, AND House} 
\label{tab:dataset} 
\begin{tabular}{cccccc}
    \hline 
Dataset&	Images&Percent.(\%) & Nest & QRcode &House \\
\hline 
Train&7506&82&4872&2143&4943\\
Val&1636&18&1087&435&1132\\

\textbf{Total}&\textbf{9142}&\textbf{100}&\textbf{5959}&\textbf{2578}&\textbf{6075}\\

\bottomrule 
\end{tabular}
\end{table}




%


\begin{figure}[!ht]
\includegraphics[width=0.5\textwidth]{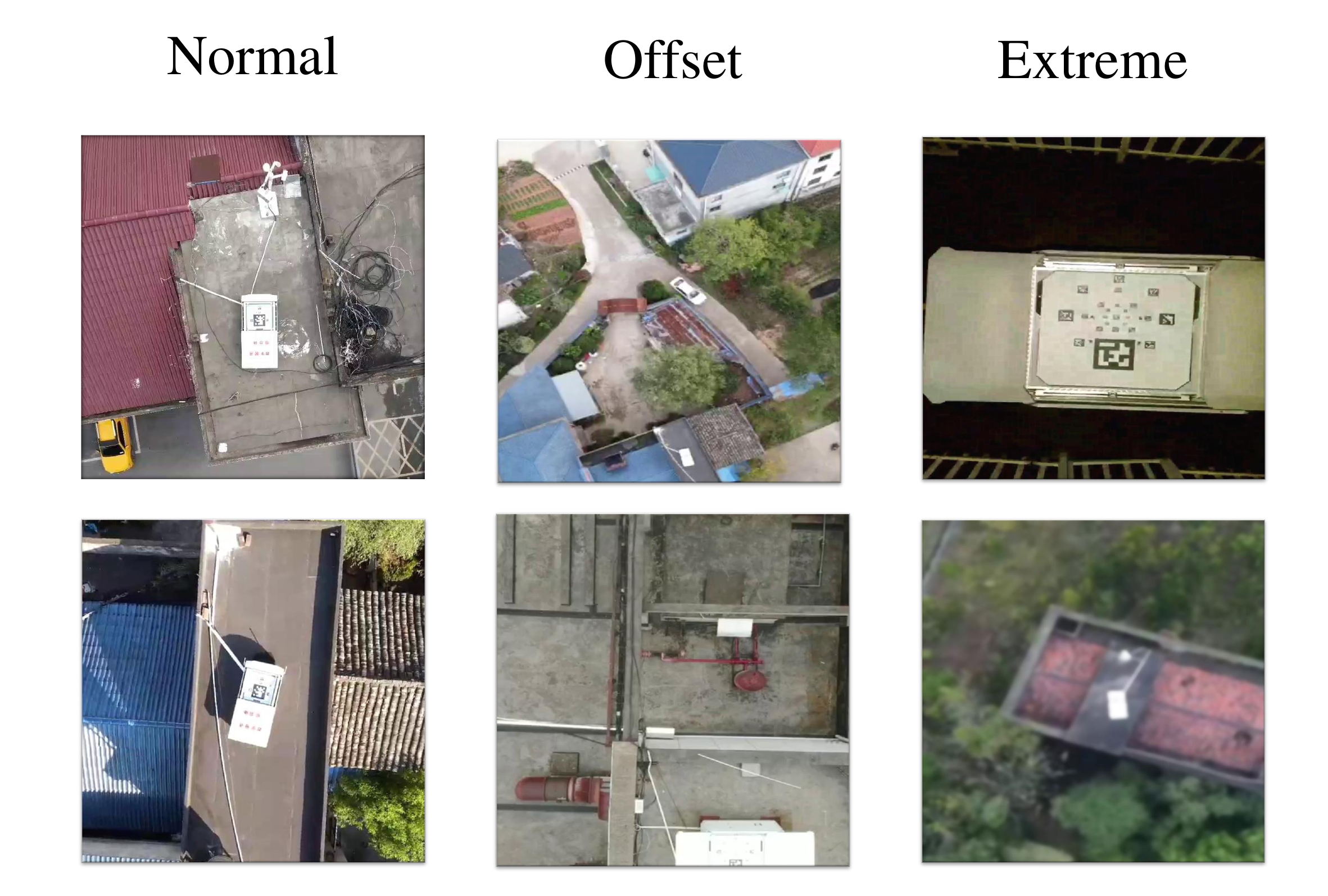}
\setlength{\abovecaptionskip}{-0.2cm}
\caption{Representative images from the UAVLandData dataset. Normal: Standard landing; Offset: Deviated landing; and Extreme: Night-time or blurred conditions.} 
\label{dataset}
\end{figure}

To train and evaluate our model, we need a large-scale, high-quality, and diverse annotated dataset. However, currently there is no publicly available dataset for UAV landing. Therefore, we collected and annotated our own dataset for UAV landing, named UAVLandData. \textcolor{black}{\autoref{dataset} gives some examples of images from UAVLandData. This
dataset has images of various lighting and weather conditions.} It includes 9,142 images covering rotated targets, small targets, and offset targets. 7,506 images are randomly selected for training and the remaining 1,636 images are used for testing. Our dataset is publicly available at \href{https://www.kaggle.com/datasets/niufaxin/uav-land-coco}{DataUAVLand}

The input size of our model is 640$\times$640. Instead of resizing the image to the target size, we crop the image at the center to 640$\times$640. As the landing procedure begins only after the UAV reaches the designated position, the critical landing zone is inherently centered in the camera's view. This enables center cropping to preserve full-resolution details of the target area while eliminating distortion from aspect ratio alteration. The cropping operation also functions as a geometric filter for deviation detection. If the target falls outside the cropped region at the start of descent, it will trigger an alert that indicates substantial positional deviation exceeding safety thresholds. This mechanism ensures undistorted target observation while establishing an automated checkpoint to abort misaligned landing attempts.

\begin{algorithm}[ht]
 \SetAlgoLined
 \KwIn{Initial labeled dataset $D_{\text{initial}}$, unlabeled dataset $D_{\text{unlabeled}}$, model $M$, stopping criterion}
 \KwOut{Labeled dataset $D_{labeled}$}
 Train model $M$ on $D_{\text{initial}}$\;
 \While{accuracy of verification set $>$ 90\%}{
   Select a sample $x^* = \arg\max_{x \in D_{\text{unlabeled}}} \frac{1}{|D_{\text{initial}}|} \sum_{(x', y') \in D_{\text{initial}}} \|x - x'\|_2$\;
   Label $x^*$\;
   Update $D_{\text{initial}} = D_{\text{initial}} \cup \{(x^*, y^*)\}$\;
   Retrain model $M$ on $D_{\text{initial}}$\;
 }
 \caption{Active Data Labeling}
 \label{Active_Learning}
\end{algorithm}

\begin{table*}[ht]
\centering
\caption{Detection results in the UAV landing.  Bold values represent the best performance.} 
\label{model_compare} 
\begin{tabular}{p{3.5cm}ccccccccc}
\hline 
\multirow{2}{*}{Method} & {Bbox}	&\multicolumn{3}{c}{Class IoU(\%)}& \multirow{2}{*}{MIoU(\%)}	& \multirow{2}{*}{params(M)} &\multirow{2}{*}{GFLOPs} & \multirow{2}{*}{FPS$_{TX2}$} & \multirow{2}{*}{FPS$_{GPU}$}	\\
\cline{3-5}
& mAP&House&	Nest	& QRcode \\
\hline 
UNet~\cite{ronneberger2015u}&-&	63.5&	40.7&	12.4&	39.2&	31.1& 345.1 & 0.5 &\textbf{243.9}\\
DeepLabV3+~\cite{chen2018encoder}&-&	92.3&	82.6&	11.1	&62.0	& 56.9 & 123.9 & 0.2&66.7\\
Mask R-CNN~\cite{he2017mask}&87.5&	85.4&	83.6&	\textbf{86.7}	&84.9 &44.5&145.4 & 1.5&89.5\\
DINO~\cite{zhang2022dino}&\textbf{91.2}&	-&	-&	-	&-	&47.0&279.3 &1.3&186.5\\
\textcolor{black}{ConvNeXt-V2}~\cite{Woo2023ConvNeXtV2}&85.0&94.3	&\textbf{88.1}	&	77.9	&\textbf{86.8}	&110.6&238.6 &0.9 &13.2\\
RTMDET-tiny-INS~\cite{lyu2022rtmdet}&87.2&	94.8&	85.2&	66.3	&82.1	&\textbf{5.6}&11.9 & 2.5 &33.4\\
AeroLite-MDNet (ours)&81.4&	\textbf{96.3}	&86.4&	76.2&	86.3&	7.2&127.7 &\textbf{7.3}&104.2\\
\hline 
\end{tabular}
\end{table*}
We employ an active learning approach to select and annotate samples, aiming to reduce labeling costs while ensuring data diversity, as shown in \autoref{Active_Learning}. First, a small subset of the data is labeled to construct an initial dataset $D_{\text{initial}}$, which will be used to train an initial model $M$. 
In each training epoch, the most diverse sample from the unlabeled dataset is selected and annotated with label $y^*$. This annotated sample is added to the initial labeled dataset, and the updated dataset is utilized to retrain the model $M$. We stop training until the model achieves an accuracy greater than 90\% in the validation set. 

\subsection{Implementation Setup}
We use the following metrics to evaluate the performance of each method: 1) IoU for each class; 2) Mean Intersection over Union (MIoU); 3) Bounding box Mean Average Precision (Bbox mAP). IoU refers to the ratio of the intersection to the union between the predicted segmentation result and the ground truth. It quantifies the overlap between the predicted and true regions. \textcolor{black}{ MIoU averages the IoU values of all classes, providing a global metric for assessing overall model performance. Unlike individual class IoUs, MIoU mitigates the dominance of certain categories, ensuring a more balanced and fair evaluation.} Bbox mAP calculates the average precision (AP) for each class, which considers both the precision and recall at various thresholds, and then takes the mean of these AP values across all classes. This metric provides a comprehensive assessment of the model's detection performance. 


In addition, we consider Giga Floating Point Operations Per Second (GFLOPs) and the number of parameters as indicators of computational complexity and memory requirements. For real-time deviation detection, GFLOPs, number of parameters, and FPS serve as crucial metrics 

We integrate our model and baselines into an end-to-end warning system for comparative experiments. In addition, AWD, Accuracy (ACC),  and False Positive Rate (FPR) are used as the evaluation metrics.



In the training process, we employ the Adaptive Moment Estimation (Adam) to update the model parameters. The initial learning rate is set to 0.001 and the weight decay is 0.0005. 
Furthermore, we employ a cosine learning schedule to reduce the learning rate. The maximum number of iterations of training is set to 300. The model is trained on a server with an NVIDIA A100-PCIE-40GB GPU and accelerated by CUDA 11.7. We employ Python 3.9 and PyTorch 1.13.1 for implementation.






\subsection{Experiment Results}

\subsubsection{Model Performance}
We compare the proposed AeroLite-MDNet with the state-of-the-art (SOTA) methods, as shown in \autoref{model_compare}. FPS$_{GPU}$ and FPS$_{TX2}$ are the FPS on the GPU and TX2 platform respectively. We select UNet~\cite{ronneberger2015u} and DeepLabV3+~\cite{chen2018encoder} as the segmentation baseline models, and DINO~\cite{zhang2022dino} as the baseline method for object detection. We also choose the classic multi-task model Mask R-CNN~\cite{he2017mask}, the state-of-the-art detection and segmentation model RTMDet~\cite{lyu2022rtmdet}, and ConvNeXt-V2~\cite{Woo2023ConvNeXtV2} as multi-task baseline models. 

Classical segmentation models, such as UNet, show limited performance, with mIoU scores of 39.2\%, respectively. DeepLabV3+ achieves a higher mIoU (62.0\%) but is constrained by its large parameter count (56.9M), making it unsuitable for embedded hardware inference. ConvNeXt-V2 achieves the highest mIoU (86.8), but its high parameter count (110.6M) limits its practicality in resource-constrained environments.

\begin{figure*}[!ht]
\includegraphics[width=7in]{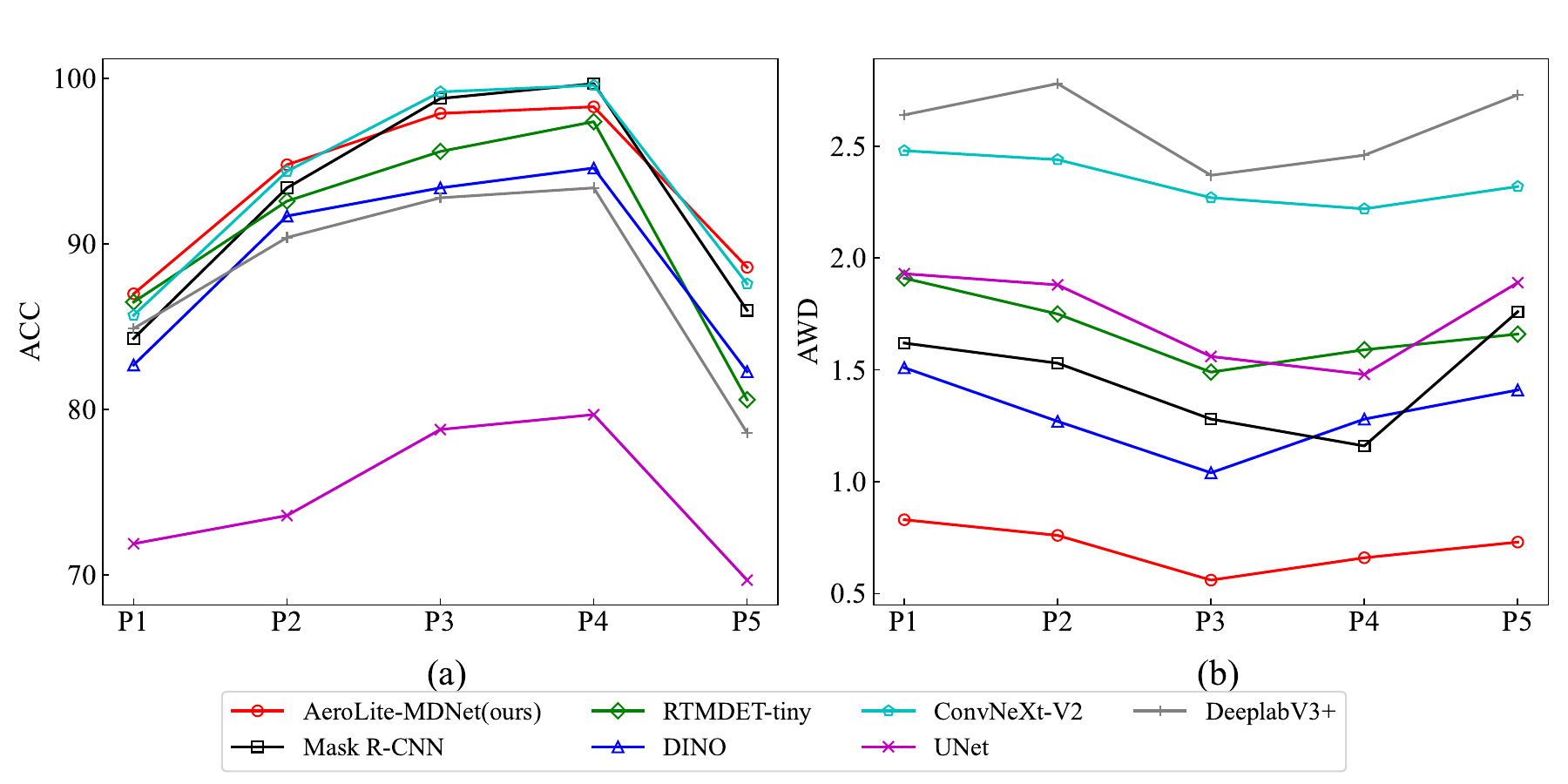}
\setlength{\abovecaptionskip}{-0.2cm}
\caption{Performance comparison of warning systems using different models during the five stages of descent (P1), stabilized descent (P2), approaching landing point (P3), posture adjustment (P4), and recovery (P5): (a) ACC, (b) AWD.} 
\label{stage_compare}
\end{figure*}

AeroLite-MDNet demonstrates a great trade-off between performance and efficiency. For QRcode, it performs well compared with other small models, such as RTMDET-tiny-INS. This indicates that our model performs well in detecting small objects. Moreover, our model achieves comparable accuracy to Mask R-CNN while having only 6 times fewer parameters. With a parameter count of only 7.2M, our model demonstrates superior compactness and efficiency compared to other models.

Considering that the model may need to work on low-performance devices in actual applications, we evaluate our method under resource-constrained conditions. Specifically, we deploy our model and other baselines on an Nvidia Jetson TX2 platform. As shown in \autoref{model_compare}, while UNet achieves the fastest inference speed on GPUs (243.9 FPS$_{GPU}$), its speed drastically drops to merely 0.5 FPS on TX2, underscoring the necessity of lightweight design. Our model maintains a high frame rate on edge devices (7.3 FPS$_{TX2}$) and efficient GPU performance (104.2 FPS$_{GPU}$), achieving a great balance between accuracy and real-time capability. 

\begin{table}[ht]
\centering
\caption{Comparison of Warning System Metrics Using Different Models. Bold values represent the best performance.
} 
\label{sys_compare} 
\begin{tabular}{c|ccc}
\hline 
Method& AWD 	&ACC(\%)& FPR(\%)\\
\hline 
UNet&1.8&77.3&6.2\\
DeepLabV3+&2.7&87.8&1.6\\
Mask R-CNN&1.4&96.7&0.7\\
DINO&1.3&93.6&1.1\\
ConvNeXt-V2&2.3	&97.2	&\textbf{0.4}\\
RTMDET-tiny-INS&1.7&95.3&0.9 \\
\hline 
AeroLite-MDNet(ours)&\textbf{0.7}&\textbf{98.6}&0.7\\
\hline 
\end{tabular}
\end{table}

\subsubsection{System Performance}

As shown in \autoref{sys_compare}, our method achieves remarkable results, with an ACC of 98.6\%, an FPR of only 0.7\%, and an AWD as low as 0.7 seconds. Compared to other methods, our approach has the lowest AWD and the highest correct warning rate. Although our false alarm rate is slightly higher than Mask R-CNNand lower than ConvNeXt-V2, both Mask R-CNN and ConvNeXt-V2 require require substantially higher time consumption, making them impractical for real-world applications.


Additionally, to show the performance at different stages of UAV landing, we divide the landing process into 5 stages, as shown in \autoref{stage_compare}, including Start descent, Stabilized descent, Approaching nest, Adjusting posture, and Recovery.
Note that as the UAV descends, the accuracy increases until the recovery stage. In the posture adjustment stage (P4), the UAV adjusts its posture to ensure the optimal angle for landing. At this point, the field of view and image reach their best state, resulting in the highest accuracy. During the recovery stage (P5), the UAV is already close to or on the nest, which causes a lack of target features in the frame and leads to a decrease in accuracy. AeroLite-MDNet demonstrates superior deviation sensitivity across all stages compared to existing models, only slightly underperforming Mask R-CNN in the stages of approaching the nest and adjusting posture.

\begin{table}[ht]
\centering
\caption{Effectiveness of SCA, CA and SEG. Bold values represent the best performance.} 
\label{Ablation Study} 
\begin{tabular}{c|c|c|c|c|c|c}
\hline 
\multirow{2}{*}{SCA} & \multirow{2}{*}{CA}& \multirow{2}{*}{SEG} &\multicolumn{4}{c}{Metrics(\%)}   \\
\cline{4-7}
&&& P&R&	F1	& mAP50:95\\
\hline 
-&-&-&85.1&80.6&82.7&70.4\\
\checkmark&-&-&84.5&84.3 &84.4&76.3\\
-&\checkmark&-&83.4&82.2&82.6&66.5\\
-&-&\checkmark&93.4&94.1 &93.5&85.6 \\
\checkmark&\checkmark&-&84.2&84.6&84.4&71.5\\
\checkmark&-&\checkmark&92.8&95.3 &93.8&83.4 \\
-&\checkmark&\checkmark&93.7&95.1 &94.3&88.2\\
\checkmark&\checkmark&\checkmark&\textbf{95.6}&\textbf{98.4}&\textbf{97.0}&\textbf{89.7}\\
\hline 
\end{tabular}
\end{table}

\begin{figure*}[!ht]
\includegraphics[width=7in]{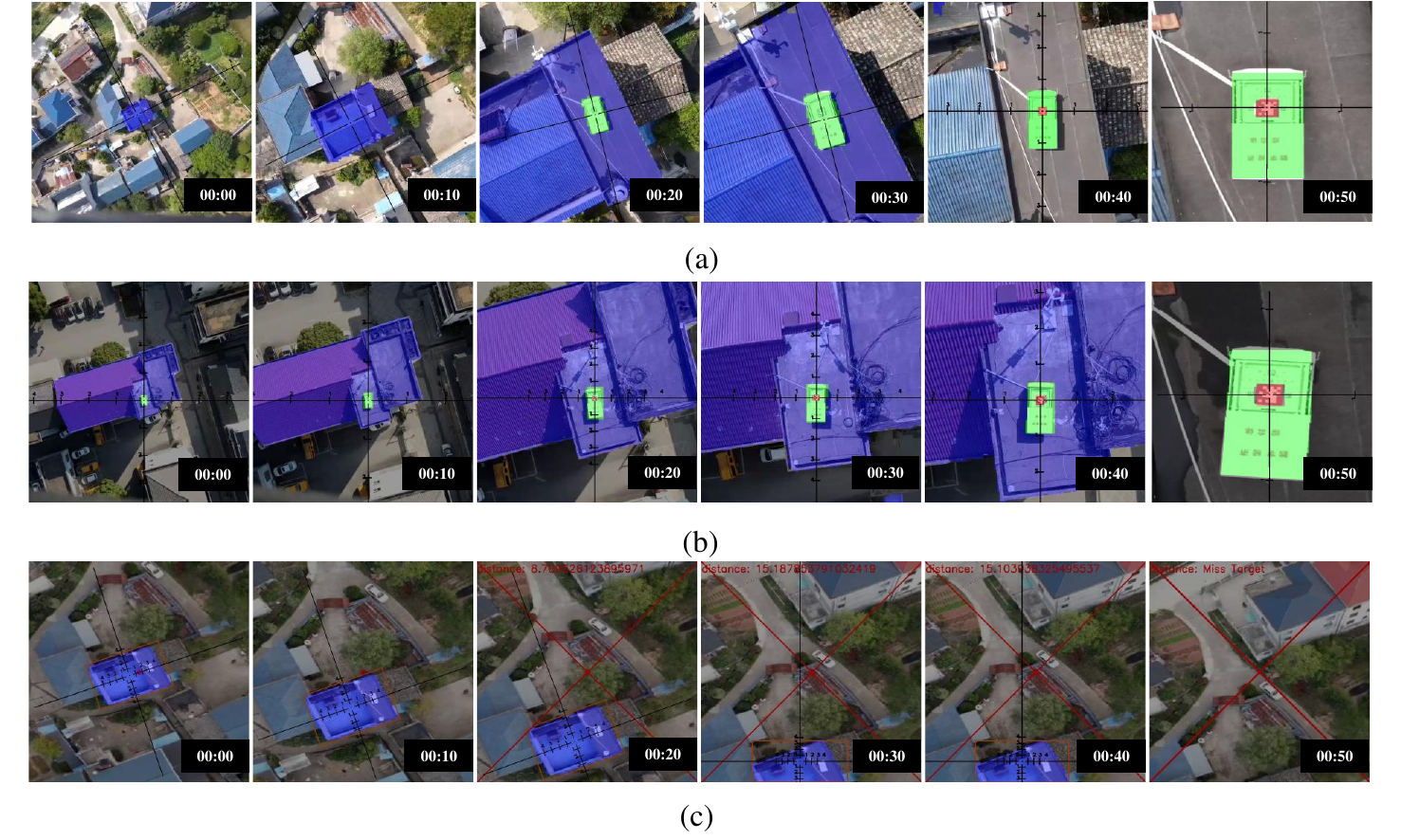}
\caption{System performance in practical applications: (a) Normal landing under adequate lighting conditions. (b) Normal landing under low-light conditions (overcast weather). (c) Offset detected under insufficient lighting conditions} 
\label{visual}
\end{figure*}

\subsection{Ablation Study}
To evaluate the contribution of the proposed SCA\_C3, CA module, and segmentation branch SEG, we conduct ablation experiments by removing each module separately. Meanwhile, we use the metrics of Precision (P), Recall (R), F1-Score (F1), and Mean Average Precision across IoU thresholds from 50\% to 95\% (mAP50:95) to evaluate the performance. Specificly, when the segmentation  branch is included, mAP is computed based on pixel-level segmentation results. In contrast, without the segmentation branch, mAP is caculated using the predicted bounding boxes.

As shown in \autoref{Ablation Study}, the baseline model does not incorporate the SCA, CA, and SEG modules, resulting in substantially lower precision, recall, F1 score, and mAP50:95 compared to our model. Notably, when the SCA module is added, recall increases substantially (from 80.6 to 84.3), while precision decreases slightly. This suggests that the model detects more targets, but may also yield more false positives. Nevertheless, the overall performance improves, particularly indicated by the rise in mAP, reflecting enhanced detection capability.

With the addition of the CA module, recall shows a slight improvement (rising from 80.6\% to 82.2\%), while precision declines (from 85.1\% to 83.4\%), and the overall mAP50:95 decreases to 66.5\%, indicating that using the CA module alone has limited effect on enhancing overall detection performance. The primary advantage of the CA module lies in its ability to perform fine-grained regulation of object detection through channel-wise attention mechanisms, which can improve detection accuracy. However, its contribution to overall performance metrics such as mAP remains relatively limited when used in isolation.

After incorporating the SEG module, all metrics show substantial improvement, especially recall (94.1) and mAP50:95 (85.6). This indicates that the segmentation module substantially improves the accuracy of object detection and localization. 

\textcolor{black}{We also conducted experiments with combinations of modules. The combination of SCA and CA did not significantly improve the model's performance. Although both channel and spatial attention were added, the limitations of CA prevented any significant improvement in metrics such as mAP50:95(71.5\%).}

\textcolor{black}{The combination of SCA and SEG improved the model's precision and recall, with noticeable improvements in mAP50:95. SCA enhanced the model's attention to key features, while SEG helped the model to more accurately segment the target, thus significantly boosting the overall performance.}

\textcolor{black}{After combining CA and SEG, the performance was further improved, especially in mAP50:95(88.2\%). The channel attention from CA and the pixel-level segmentation from SEG complemented each other, further enhancing the model's ability to focus on important features and target boundaries, resulting in better overall performance.}

\textcolor{black}{Finally, when the SCA, CA, and SEG modules are added simultaneously, precision (95.6\%), recall (98.4\%), F1 score (97.0\%), and mAP50:95 (89.7\%) all reach their highest values, indicating that these modules, through the combination of spatial and channel attention mechanisms and pixel-level segmentation, significantly enhance the model's ability to focus on key features.}

The system validation in various real-world scenarios is shown in \autoref{visual}. In \autoref{visual}(a), a normal landing process under adequate lighting conditions. The target area, nest, and QR code are clearly visible during the landing process, and the model recognizes the target accurately. The crosshair in the image indicates the target position. \autoref{visual}(b) shows the landing process under low-light conditions. In this scenario, due to rain and overcast weather, dark-colored rooftops and ground surfaces reduce visual contrast and increase detection difficulty. Nevertheless, the system is still able to reliably detect the landing nest. The white nest maintains a strong contrast with the darker background under low-light conditions, allowing the model to accurately identify the landing target. \autoref{visual}(c) shows an offset landing under similar low-light conditions. When the UAV deviates from the designated landing position beyond a predefined threshold, the system automatically issues a warning. This indicates that our system performs robustly in challenging environments.

\section{Discussion}
The proposed method performs well in monitoring UAV landings in various environments. However, the current system mainly relies on RGB cameras, additional spotlights are required for illumination at night, increasing system complexity and energy consumption. Future research could consider integrating multiple types of sensors, such as infrared cameras, LiDAR, or millimeter-wave radar, to enhance perception in various environments and improve overall robustness and adaptability. Moreover, the current approach focuses on monitoring, with limited integration into the UAV control system. Future work may explore integration, such as jointly optimizing perception and control. This would enable more efficient coordination and response during the landing process, further improving landing accuracy and stability.
\section{Conclusion}
\label{sec:con}
This paper proposes a practical end-to-end method for real-time UAV landing warning. To the best of our knowledge, our work is the first for detecting  UAV deviations in real time. Our method is evaluated in real UAV landing scenarios and achieves an overall MIoU of 86\% for 3 typical targets. The results demonstrate that most targets can be effectively identified and located, which enables effective warning in cases of GPS anomalies and failures, reducing losses caused by UAV landing deviations. Additionally, our experiments show that the end-to-end inference of our system on a Jetson TX2 is 7.3 FPS. Such speed meets the real-time warning requirements for UAV landing and has low hardware requirements. Our future work will focus on reducing the processing time of target recognition and enhancing the robustness of the system. 

\bibliographystyle{Bibliography/IEEEtranTIE}
\bibliography{Bibliography/ref}\ 

\begin{IEEEbiography}[{\includegraphics[width=1in,height=1.25in,clip,keepaspectratio]{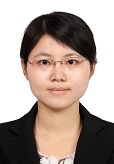}}] {Haiping Yang} received the Ph.D. degree in cartography and geographic information system from the Institute of Remote Sensing and Digital Earth, Chinese Academy of Sciences, Beijing, China, in 2015. She is currently an Assistant Professor with Zhejiang University of Technology. Her research interests include high spatial resolution remote sensing image classification, change detection, and neural networks.
\end{IEEEbiography}

\begin{IEEEbiography}[{\includegraphics[width=1in,height=1.25in,clip,keepaspectratio]{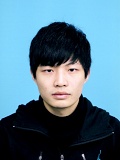}}] {Huaxing Liu} received the B.S. degree in marine technology from Zhejiang Ocean University, Zhoushan, China, in 2018. He has been pursuing the M.S. degree in computer technology at Zhejiang University of Technology, Hangzhou, China, since 2022. His current research interests include intelligent inspection methods for unmanned aerial vehicles (UAVs).
\end{IEEEbiography}

\begin{IEEEbiography}[{\includegraphics[width=1in,height=1.25in,clip,keepaspectratio]{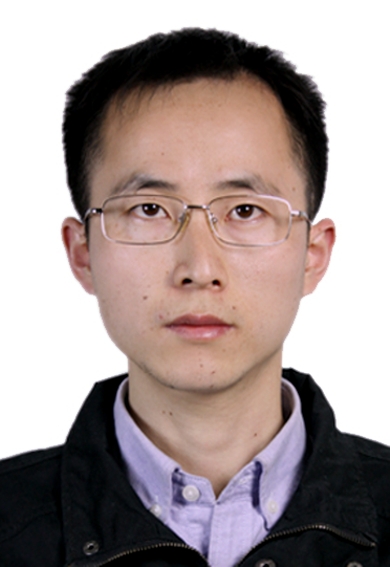}}] {Wei Wu} received the B.E. degree in land resource management from Anhui Normal University, Wuhu, China, in 2007, and the Ph.D. degree in cartography and geographic information system from the University of Chinese Academy of Sciences, Beijing, China, in 2013. He is currently an Associate Professor with the School of Computer Science and Technology, Zhejiang University of Technology, Hangzhou, China. His research interests include remote sensing information extraction.
\end{IEEEbiography}

\begin{IEEEbiography}[{\includegraphics[width=1in,height=1.25in,clip,keepaspectratio]{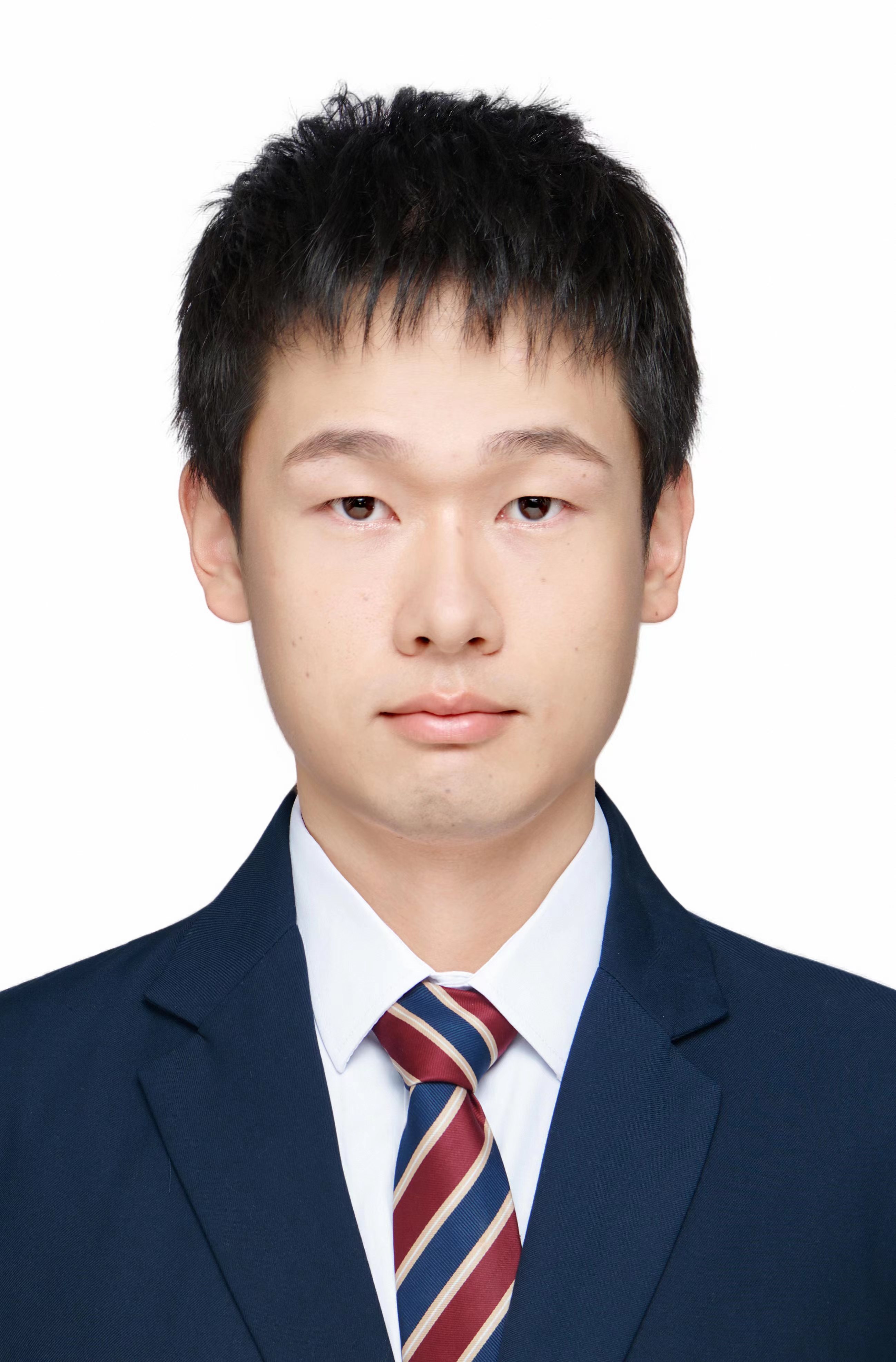}}] {Zuohui Chen} received the B.E. degree in automation from the Zhejiang University of Technology, Hangzhou, China, in 2019, and the Ph.D. degree in control theory and engineering from the Zhejiang University of Technology, in 2024. He is currently a Post-Doctoral Research Fellow with the College of Geoinformatics, Zhejiang University of Technology. His current research interests include trustworthy AI applications and remote sensing.
\end{IEEEbiography}

\begin{IEEEbiography}[{\includegraphics[width=1in,height=1.25in,clip,keepaspectratio]{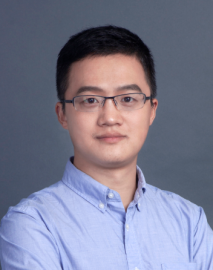}}] {Wu Ning} received the B.E. degree in urban planning from the Zhejiang University, Hangzhou, China in 2008 and the Ph.D. degree in architectural design and theory in 2013. He is the founder and CEO of Quzhou Southeast Feishi Technology Co., Ltd., part-time vice president of Southeast Digital Economy Development Research Institute, president of Longyou Branch of Southeast Digital Economy Development Research Institute, director of Digital Space Center of Southeast Digital Economy Development Research Institute, associate professor of Quzhou University. His research interests include application of spatial digital technology.
\end{IEEEbiography}

\end{document}